\documentclass[conference]{IEEEtran}
\IEEEoverridecommandlockouts

\usepackage[dvipsnames, table]{xcolor}
\usepackage{pgf}
\usepackage{pgfplots}
\pgfplotsset{compat=1.14}
\usepackage{float}
\usepackage{booktabs}
\usepackage{siunitx}
\usepackage{cite}
\usepackage{bbm}
\DeclareSIUnit{\nothing}{\relax}
\usepackage{multirow}
\usepackage{nicefrac}
\usepackage[hidelinks]{hyperref}
\usepackage{mathtools}

\usepackage[absolute,showboxes]{textpos}

\TPMargin{5pt}

\newcommand{\copyrightstatement}{
    \begin{textblock}{0.84}(0.08,0.93)
         \noindent
         \footnotesize \textcopyright 2024 IEEE. Personal use of this material is permitted.  Permission from IEEE must be obtained for all other uses, in any current or future media, including reprinting/republishing this material for advertising or promotional purposes, creating new collective works, for resale or redistribution to servers or lists, or reuse of any copyrighted component of this work in other works.
    \end{textblock}
}

\DeclareSIUnit{\nothing}{\relax}
\DeclareSIUnit\mac{MAC}
\usepackage{tikz}
\usepackage{textcomp}
\usepackage[doipre={DOI:~}]{uri}
\usepackage{lipsum}

\definecolor{somegray}{rgb}{0.5, 0.5, 0.5}
\newcommand{\darkgrayed}[1]{\textcolor{somegray}{#1}}
\makeatletter
\newcommand*\titleheader[1]{\gdef\@titleheader{#1}}
\AtBeginDocument{%
  \let\st@red@title\@title
  \def\@title{%
    \vskip-1.4em
    \bgroup\normalfont\large\centering\@titleheader\par\egroup
    \vskip0.0em\st@red@title}
}

\makeatother

\titleheader{\darkgrayed{This paper has been accepted for publication in the DATE 2024 conference\\\copyright 2024 IEEE.}}

\title{Adaptive Deep Learning for Efficient Visual Pose Estimation aboard Ultra-low-power Nano-drones\\
\thanks{This publication is part of the project PNRR-NGEU which has received funding from the
MUR – DM 118/2023.}}

\author{\IEEEauthorblockN{Beatrice Alessandra Motetti\IEEEauthorrefmark{1}, Luca Crupi\IEEEauthorrefmark{2}, Mustafa Omer Mohammed Elamin Elshaigi\IEEEauthorrefmark{1}, Matteo Risso\IEEEauthorrefmark{1},\\ Daniele Jahier Pagliari\IEEEauthorrefmark{1}, Daniele Palossi\IEEEauthorrefmark{2}\IEEEauthorrefmark{3}, Alessio Burrello\IEEEauthorrefmark{1}}

\IEEEauthorblockA{ 
\IEEEauthorrefmark{1} Politecnico di Torino, Turin, 10129, Italy\\
\IEEEauthorrefmark{2} Dalle Molle Institute for Artificial Intelligence, USI and SUPSI, Lugano, 6962, Switzerland\\
\IEEEauthorrefmark{3} Integrated Systems Laboratory (IIS), ETH Z\"urich, Zurich, 8092, Switzerland}

\IEEEauthorblockA{Emails: name.surname@polito.it, name.surname@idsia.ch} 
}

\begin{document}

\bstctlcite{IEEEexample:BSTcontrol}

\maketitle
\copyrightstatement

\begin{abstract}

Sub-\SI{10}{\centi\meter} diameter nano-drones are gaining momentum thanks to their applicability in scenarios prevented to bigger flying drones, such as in narrow environments and close to humans.
However, their tiny form factor also brings their major drawback: ultra-constrained memory and processors for the onboard execution of their perception pipelines.
Therefore, lightweight deep learning-based approaches are becoming increasingly popular, stressing how computational efficiency and energy-saving are paramount as they can make the difference between a fully working closed-loop system and a failing one.
In this work, to maximize the exploitation of the ultra-limited resources aboard nano-drones, we present a novel adaptive deep learning-based mechanism for the efficient execution of a vision-based human pose estimation task.
We leverage two State-of-the-Art (SoA) convolutional neural networks (CNNs) with different regression performance vs. computational costs trade-offs.
By combining these CNNs with three novel adaptation strategies based on the output's temporal consistency and on auxiliary tasks to swap the CNN being executed proactively, we present six different systems.
On a real-world dataset and the actual nano-drone hardware, our best-performing system, compared to executing only the bigger and most accurate SoA model, shows 28\% latency reduction while keeping the same mean absolute error (MAE), 3\% MAE reduction while being iso-latency, and the absolute peak performance, i.e., 6\% better than SoA model.
\end{abstract}

\section{Introduction}
\label{sec:intro}
\looseness=-1
Nano-sized Unmanned Aerial Vehicles (UAVs) have emerged as a groundbreaking technology thanks to their under-\SI{10}{\centi\meter} size and less than \SI{40}{\gram} weight. They have opened new frontiers for applications in GPS-denied environments, confined spaces, and close to humans~\cite{pulp-dronet}.
However, their miniaturization comes at the price of simplified equipped sensory, mechanical, and computational subsystems.
The power available for computing, typically less than 100 milliwatts, is very limited compared to standard-sized commercial drones equipped with powerful CPUs or even mobile GPUs.
Consequently, nano-UAVs typically mount Microcontroller unit (MCU)-class processors.
This poses a significant challenge when executing resource-intensive perception, planning, and control algorithms. 
At the same time, offloading these tasks to a remote server through a continuous stream of raw sensor data is restrictive given the instability of the wireless connection and its limited range and does not guarantee a predictable response latency~\cite{varghese2016challenges}.

Therefore, to enable the onboard intelligence on nano-UAVs, lightweight Convolutional Neural Networks (CNNs) offer a possible solution to address these challenges in perception tasks.
They have been shown to work effectively even with minimal sensor data, accommodating ultra-low-power cameras with limited resolution and dynamic range~\cite{tiny-dronet, frontnet}.
Furthermore, they offer predictable and fixed computational and memory requirements during inference, enabling real-time operation without wireless coverage for offloading.

So far, the State-of-the-Art (SoA) autonomous navigation of nano-UAVs is characterized by deploying one static CNN~\cite{frontnet, cereda_deep_2023}, acting as a high-level perception module.
If, on the one hand, this approach offers a viable solution, on the other hand, it limits the level of accuracy and computational efficiency of the onboard perception.
For some deployment scenarios, this design choice means addressing a simple task with a too-complex CNN -- wasting computational resources and, therefore, energy -- while in other cases, a static CNN might underperform due to a too-complex environment.

Recently, adaptive machine learning has gained traction as a method to dynamically control the trade-off between accuracy and complexity in neural network-based applications~\cite{Park2015}.
To cope with variable-complexity inputs, adaptive approaches build \textit{ensembles of models} that are conditionally executed based on environmental conditions: in the simplest embodiment, a more lightweight, less memory demanding, and faster model is employed when a \textit{task-specific policy} function deems the processed datum to be relatively easy, while a larger, more powerful model is used for more challenging inferences.

This work uses this paradigm to demonstrate the potential for optimizing a vision-based human pose estimation task onboard a nano-UAV: the fundamental algorithmic building block for ``people monitoring'' and ``follow-me'' applications.
To the best of our knowledge, our work proposes the first employment of an adaptive system capable of adjusting its complexity based on the mission's environment.
Our detailed contribution is as follows:
\begin{itemize}
\item We propose two distinct adaptive ensembles, which combine three models from~\cite{cereda_deep_2023}. The two reported less accurate networks are used as \textit{small} models in our ensembles, and the most accurate network as \textit{big} model in both ensembles. Further, we propose three novel adaptation policies based on the output's temporal consistency of pose estimation and on auxiliary tasks explicitly tailored for visual human pose estimation aboard nano-UAVs.
\item We demonstrate that our adaptive algorithms, i.e., the ensembles combined with our proposed policies, outperform the SoA results on two datasets for the same robotic task.
\item We demonstrate that, compared to SoA static CNNs, our approach generates a broader solution space with numerous intermediate points that provide a good trade-off between accuracy and complexity compared to static models.
Compared to the most accurate SoA network, we reduce the Mean Absolute Error (MAE) on the estimated pose by 3.15\% while maintaining a constant latency.
When we consider iso-MAE performance, we instead reduce the latency of execution by 28.03\%. 
\end{itemize}
\section{Background \& Related Work}
\label{sec:background_related}

\subsection{Human Pose Estimation aboard Nano-Drones}
Our work is in the context of drone navigation tasks aimed at following a human subject, maintaining a constant distance~\cite{frontnet}. 
The input data are grey-scale images captured by an onboard camera, while the outputs are poses then converted into set-points for the low-level control.
More specifically, we focus on the perception sub-task, which consists of a relative pose estimation between a human subject and a nano-UAV, approached using a CNN model.
While SoA computer vision techniques~\cite{Sun_2018_ECCV,luvizon18multitask,dense-pose} have achieved remarkable accuracy on this problem, they remain beyond the practical capabilities of nano-drones due to their high computational demands ($\sim10^{10}$ Multiply and ACcumulate - MAC - operations)~\cite{dense-pose}.
In contrast to these algorithms, which estimate complete skeletal structures~\cite{luvizon18multitask} or dense 3D mesh representations of individuals~\cite{dense-pose}, CNNs deployed on nano-drones focus their predictions on a minimal set of elements: the head position in 3D space ($x, y, z$) and a rotation angle relative to the gravity z-axis ($\phi$). 

However, deploying CNNs on nano-UAVs still presents substantial challenges due to the limited hardware resources, as mentioned in Sec.~\ref{sec:intro}.
Traditional solutions relying on off-the-shelf MCUs, such as those in~\cite{neuralswarm,uwbbias}, can only accommodate simple neural architectures to achieve real-time performance.
For example, the network in~\cite{uwbbias} performs a modest \SI{27}{\kilo\mac} per frame at a rate of \SI{100}{\hertz}. 
Consequently, these approaches are primarily suitable for processing low-dimensional input signals and ill-suited for the computational demands of camera image processing.
Thus far, the PULP-Frontnet CNN and its optimized variants~\cite{frontnet, cereda_deep_2023} stand as one of the few instances where real-time visual-based human pose estimation tasks have been successfully executed aboard a nano-UAV.
These methods leverage general-purpose ultra-low-power multi-core System-on-Chip (SoC) architectures, tailored software deployment pipelines, and fine-tuned CNN architectures to carry out the task efficiently.
PULP-Frontnet has a complexity of \SI{14.7}{\mega\mac} and achieves an inference rate of \SI{45}{\hertz} with a power consumption of \SI{92.2}{\milli\watt} when deployed on a Crazyflie~2.1 nano-drone~\cite{frontnet}.
Subsequently, in~\cite{cereda_deep_2023}, Neural Architecture Search (NAS) was employed to further statically optimize this CNN, obtaining a latency reduction of 13.2\%~\cite{cereda_deep_2023} without accuracy degradation.

All advancements in executing CNN-based vision tasks on nano-drones have focused on \textit{static} (data-independent) optimizations. 
This approach misses an important opportunity since, in typical drone navigation tasks, the environment can often change, e.g., from indoor to outdoor, from empty spaces to crowded ones, etc, calling for a perception method that can adapt to changes in the input data distribution. 
This adaptability can lead to a faster inference for ``easy'' input data, which provides a higher average throughput and, consequently, a faster response from the control part. 

\subsection{Adaptive Machine Learning}
The challenge of optimizing Deep Learning (DL) models for execution on ultra-low-power edge nodes, trading off slight reductions in accuracy for substantial gains in latency, energy efficiency, or memory usage, has been the subject of extensive research recently~\cite{Sze2017,pagliari_plinio_2023}. 
One significant categorization of DL optimization techniques distinguishes them as \textit{static} and \textit{dynamic}~\cite{Daghero2021energy}.  
The former involves compressing a model before deployment, either during the training phase or post-training.
Static methods include quantization, pruning, and NAS~\cite{pagliari_plinio_2023}, now commonly used to optimize neural networks.
A critical limitation of these approaches is their inability to adapt the inference complexity at runtime when environmental conditions evolve.

Dynamic (or ``adaptive'') inference techniques, including the approaches presented in this paper, aim to address this limitation. 
An adaptive inference's complexity is reduced when the input data is ``simple'', and a lower amount of computation suffices to ensure high accuracy (e.g., for a vision task, the relevant object is well-centered in the frame, not blurred, and against a uniform background).
Conversely, more compute budget is allocated for ``challenging'' inputs (e.g., a moving or partially obstructed object)~\cite{Park2015, Tann2016, har_journal,teerapittayanonBranchyNet2016}.
The easiest way to implement an adaptive system is by appropriately combining the outputs of two independent static models~\cite{Park2015}: a smaller, less complex model for easy-to-process inputs and a larger, more accurate one for the others. 
Other alternatives also exist, such as using only parts of the channels of a complex model to obtain the lightweight one~\cite{Tann2016,har_journal}, or early-exiting after a subset of layers~\cite{teerapittayanonBranchyNet2016}.

Regardless of the specific scheme, a crucial component for all adaptive DL systems is a discrimination \textit{policy}, which determines which model to activate for a given input data~\cite{Park2015}. 
An effective policy should provide an acceptable loss of accuracy compared to the most complex model while resulting in a substantial computational saving.
Notably, while policies for classification tasks have been extensively explored, the same is not true for regression tasks.
To the best of our knowledge, we are the first to propose two novel policies that consider the temporal correlation between images captured by a camera and an auxiliary task to gauge the input's complexity. 
In Sec.~\ref{sec:methods}, we delve into a detailed analysis of these policies.
\section{Materials \& Methods} \label{sec:methods}

\subsection{Static Neural Networks}
Inspired by~\cite{Park2015}, we build our adaptive system from an ensemble of two independent ``static'' CNNs of different complexity and accuracy.
Namely, we consider two ensembles by combining three SoA models proposed in~\cite{cereda_deep_2023}, whose MAE, number of parameters, and NMACs are reported in Tab.~\ref{tab:static_models}.
F$^1$ and F$^2$ are two versions of a convolutional network based on PULP-Frontnet composed by 7 convolutional layers, the first with a $5\times5$ filter and the others with a $3\times3$ filter.
The difference between the two networks is in the number of filters (or channels) of each layer.
M$^{1.0}$ is a reduced version of a classical MobileNet v1.0, pruned by a SoA NAS algorithm~\cite{pagliari_plinio_2023} to minimize the number of filters.
All networks have been trained on the open source dataset of~\cite{cereda_deep_2023}.
For our experiments, we combine these three CNNs in pairs to create two different ensembles, fixing the big model to be M$^{1.0}$ and changing the small model between the other two.

\begin{table}[t]
    \begin{center}
      \caption{Static models metrics extracted from~\cite{cereda_deep_2023}.}
      \label{tab:static_models}
      \renewcommand{\arraystretch}{1.25}
      \footnotesize
      \begin{tabular}{lccccccc}
        \hline
        \textbf{Network} & \multicolumn{5}{c}{\textbf{MAE}} & \textbf{Params} & \textbf{MAC} \\
         & $x$ & $y$ & $z$ & ${\phi}$ & sum & & \\\hline
    \textbf{F$^1$} &0.27& 0.27& 0.28& 0.52& 1.34& 14.8 k& 4.51 M \\
        \textbf{F$^2$} &  0.21& 0.18& 0.24& 0.46&  1.10 &44.5 k& 7.09 M\\
        \textbf{M$^{1.0}$} &0.19& 0.14& 0.23& 0.48& 1.04 &46.8 k&11.42 M\\  
        \hline
      \end{tabular}
    \end{center}
    \vspace{-0.2cm}
  \end{table}

\subsection{Adaptive Inference for Visual Pose Estimation}
Adaptive inference for regression problems must be formulated differently from the case of classification. 
Most adaptive policies rely on the class probabilities estimated by a softmax classifier function to compute ``confidence'' scores~\cite{Daghero2021energy, Park2015, Tann2016,har_journal,teerapittayanonBranchyNet2016}. 
This cannot be directly extended to a regressor, which produces a pointwise estimate of continuous variables (in our case, $x$, $y$, $z$, and $\phi$) without associated confidence. 
Given this limitation, we introduce ad-hoc policies to manage inference adaptively based on our specific problem's characteristics.
Specifically, we introduce two novel policies, \textit{Output-based Partitioning} and \textit{Auxiliary Task-based Partitioning}, to realize adaptive inference for visual pose estimation on nano-drones.

Note that an \textit{ideal} policy should execute the big model (M$^{1.0}$) if and only if it provides a lower prediction error than the small model on the currently considered input.
At the same time, the policy itself should have negligible computational complexity compared to the execution of the two models to avoid nullifying the gains.

\subsubsection{Output-based Partitioning}

\begin{figure}[t]
    \centering
    \includegraphics[width=0.65\columnwidth]{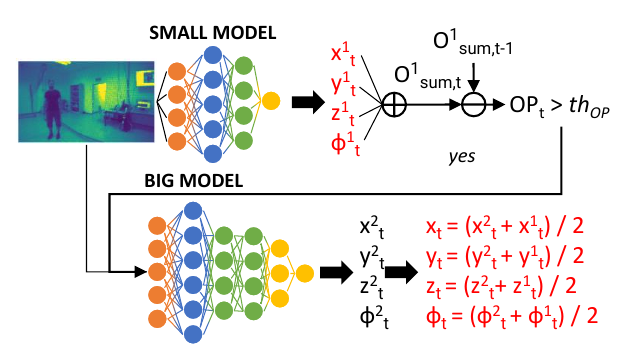}
    \caption{OP policy: the small model is always executed to compute a first set of outputs, that are possibly corrected by the big model if they differ significantly from the prediction at the previous time stamp.}
    \label{fig:out_metric}
    \vspace{-0.1cm}
\end{figure}

The Output-based Partitioning (\textit{OP}) policy is based on assessing changes in predictions across consecutive time steps. 
As illustrated in Fig.~\ref{fig:out_metric}, at each new time stamp $t$, we first run the small model and monitor variations,  compared to time $t-1$, in the predicted, min-max scaled (thus dimensionless) $x$, $y$, $z$, and $\phi$ values. Specifically, we compute:
\begin{equation}
    O_{\text{sum}} = x + y + z + \phi,\ \ OP_{t} = O_{\text{sum},t} - O_{\text{sum},t-1}
\end{equation}
A higher $OP_{t}$ score indicates that the relative position between human and drone is changing rapidly, necessitating the drone to change its location.
This is associated with more challenging predictions due to the moving camera and the subject's distance from the center of the image.
Conversely, a lower $OP_{t}$ signifies a slowly moving or stationary drone.

Based on these considerations, we build an adaptive system that works as follows: 
when $OP_{t} \le th_{OP}$, the predictions of the small model are used ``as is''.
Vice versa, the \textit{big} network is executed when $OP_{t} > th_{OP}$, and the final outputs are computed as the \textit{average} between the small and big model predictions, based on the observation that using both models concurrently, as an ensemble, leads to enhanced accuracy.
Notably, $th_{OP}$ is a tunable threshold that can be set at runtime to manipulate the prediction error vs. complexity trade-off (lower $th$ implies more big model executions, and vice versa).

\begin{figure}[t]
    \centering
    \includegraphics[width=.8\columnwidth]{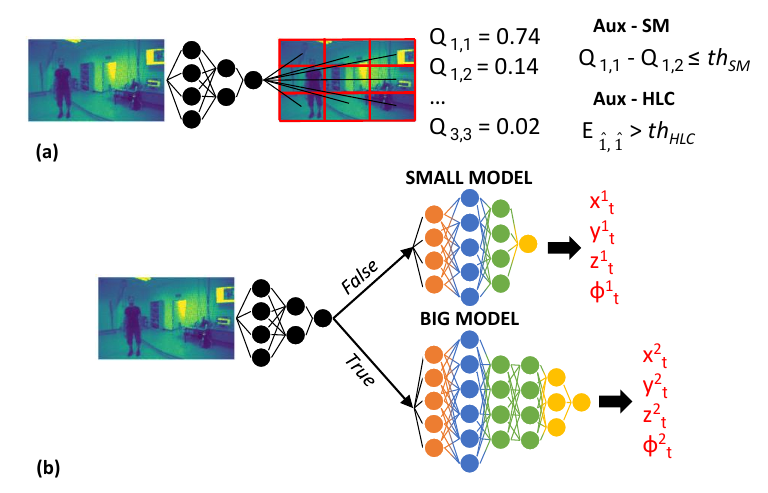}
    \caption{Auxiliary task-based partitioning. (a) A first auxiliary CNN is executed to localize the head, and one of the two policies is applied. (b) Based on the policy outcome, one of the two models of the ensemble is executed.}
    \label{fig:aux-task}
    \vspace{-0.1cm}
\end{figure}

The critical limitation of this scheme is that the small model is always executed since its outputs are needed to compute $OP_{t}$.
Numerically, the total cost is:

\begin{equation}
C = C_{small} + N_{big}(th_{OP}) \cdot C_{big}
\end{equation}
where $C_{small}$/$C_{big}$ are the NMACs or latency of the small/big models, respectively, and $N_{big}(th_{OP}) = \sum_{t} \mathbbm{1}(OP_t > th_{OP})$, with $\mathbbm{1}(\cdot)$ being the indicator function, is the number of times the big model is called.

\subsubsection{Auxiliary Task-based Partitioning}

To avoid the constant execution of the small model, we propose an alternative policy that leverages an auxiliary classification task executed before the pose estimation.
As shown in Fig.~\ref{fig:aux-task}, we train a small additional CNN to locate the human head in one of the grid cells superimposed to the input image.
We test grids of dimensions $2\times 2$, $3\times 3$, and $8\times 6$.

We build the auxiliary CNN as a strongly reduced version of PULP-Frontnet, with additional stride-2 pooling layers to rapidly shrink the activation tensors.
We start from 4 convolution+pooling blocks with 8, 16, 32, and 64 filters, respectively, and a final fully connected layer for just 656 kMACs.
We then prune unimportant filters from this network using the mask-based method from~\cite{pagliari_plinio_2023}.

Formulating head localization as a multi-class classification allows us to derive two adaptive inference policies. 
The first uses the SoA score margin method for confidence estimation (Aux-SM), whereas the second is based on domain knowledge and is called Head Localization-Class (Aux-HLC).

\textbf{Aux-SM} assumes that if the auxiliary CNN's confidence is low in classifying the head position, then the original pose regression task will be harder for a given input (e.g., because the image is blurry). 
We estimate the auxiliary task confidence with the score margin:

\begin{equation}
SM_t = \max(Q_{i,j,t}) - \mathrm{second\_max}(Q_{i,j,t})
\end{equation}

where $Q_{i,j,t}$ represents the probabilities assigned to the $(i, j)$ grid cell for frame $t$. 
Higher confidence values indicate that the prediction is more reliable; therefore, the input image is easier, and pose estimation can be done with the small model. 
In this case, we define a tunable threshold $th_{SM}$ and invoke the big model if and only if $SM_t \le th_{SM}$.

\begin{figure}
    \centering
    \includegraphics[width=0.67\columnwidth]{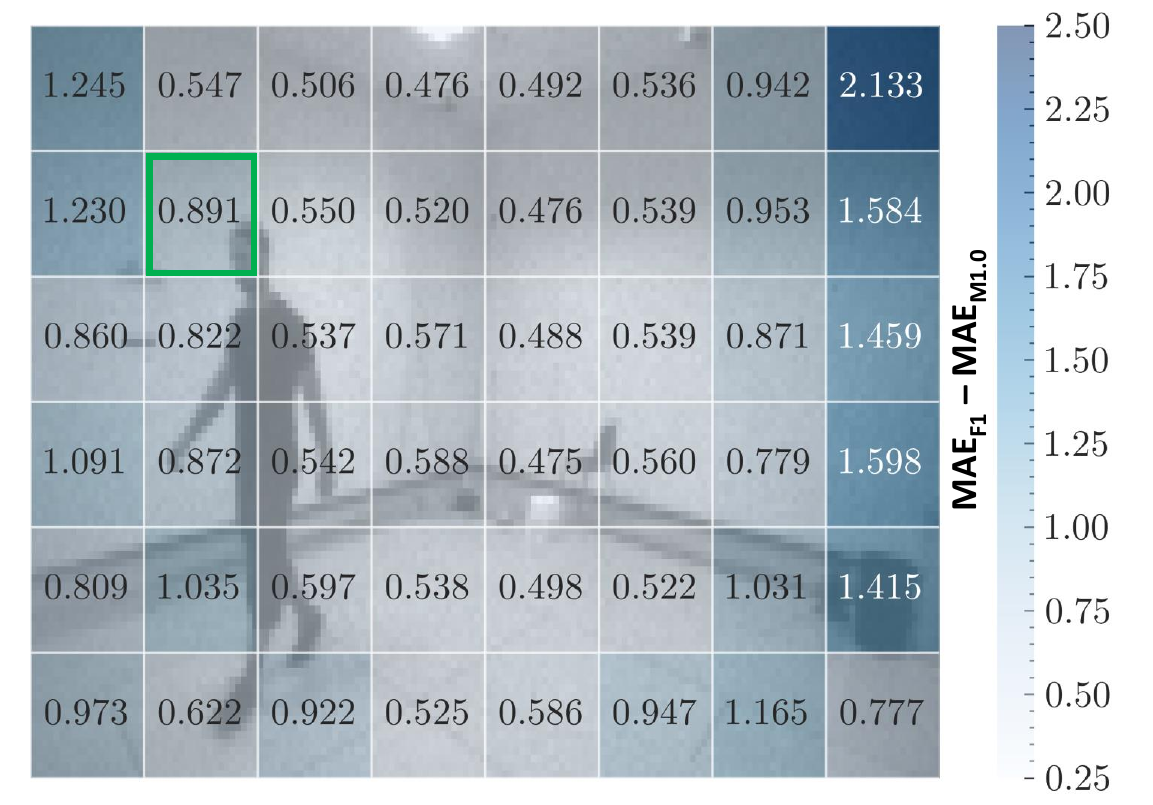}
    \caption{$8\times$6 grid division of the input images. In each quadrant, the difference between the MAE of F$^1$ and M$^{1.0}$ is reported. A green square marks the quadrant to which the head of the person belongs.}
    \label{fig:pretext_imshow}
    \vspace{-0.1cm}
\end{figure}

\textbf{Aux-HLC} assumes that pose estimation is harder when the head is in specific locations of the image (i.e., borders/corners vs center). 
To create a policy from this hypothesis, we rely on a validation set and create an error map, which indicates how much the big model outperforms the small one in each grid cell. 
Namely, each error map location contains $E_{i,j} = MAE_{small,i,j} - MAE_{big,i,j}$, that is, the average MAE difference computed on all validation samples for which the ground truth head location is in cell $(i,j)$.
The policy invokes the big model if and only if $E_{\hat{i},\hat{j}} > th_{HLC}$, where $(\hat{i},\hat{j})$ is the grid cell predicted by the auxiliary CNN at time $t$.
Fig.~\ref{fig:pretext_imshow} shows the error map for models F$^1$ and M$^{1.0}$.
As shown, the difference increases at the edges and even more at the corners.

For both \textit{Aux} policies, the total cost is:
\begin{equation}
    C = C_{Aux} + (1 - N_{big}(th))\cdot C_{small} + N_{big}(th) \cdot C_{big}
\end{equation}
where $th \in \{th_{SM}$, $th_{HLC} \}$ and depends on the policy, and $C_{Aux}$ is the cost of the auxiliary CNN. 
Given that the latter is lower than $C_{small}$, it is evident that these policies incur a lower total cost compared to the OP policy when the big model is invoked frequently. 
However, with the auxiliary policies, the small and big model outputs are not ensembled for hard inputs, possibly leading to a higher error.

\subsection{Target Platform} \label{sec:system_design}

The target robotic platform is the Crazyflie 2.1, a nano-drone weighting only \SI{27}{\gram} produced by Bitcraze. 
The nano-drone mounts an AI-deck board featuring a camera and the GAP8 SoC by GreenWaves Technologies~\cite{flamand2018gap}. 
The camera acquires grayscale images up to 320x320 pixels, while the SoC extends the computational capabilities offered by the onboard STM32 processor for control purposes.
GAP8 is a parallel ultra-low power platform~\cite{conti2017iot} with hardware support for integer-only arithmetic. 
It features a single-core fabric controller (FC) and an 8-core cluster (CL).
The FC orchestrates transfers between memories and offloads computationally intense tasks to the CL.
The SoC also includes a \SI{64}{\kilo\byte} L1 memory shared between the CL cores and a \SI{512}{\kilo\byte} L2 memory within the FC.
The off-chip memories consist of an 8 MB DRAM and a 64 MB FLASH. 
The transfers between DRAM and L2 are operated autonomously through a programmable micro-DMA.
Transfers between L2 and L1 are performed by a DMA that is in the CL domain.

To deploy CNNs on this platform, we exploit two tools: PLiNIO~\cite{pagliari_plinio_2023} and DORY~\cite{burrello2020dory}.
PLiNIO is a multi-optimization library that we leverage for shrinking the auxiliary CNN channels, as well as for applying Quantization-Aware Training (QAT) to all models to obtain integer-only implementations at 8-bit precision deployable on GAP8.
The DORY compiler automatically converts the trained integer models from Python to C code, leveraging an optimized kernel library. 
It applies hardware-aware tiling and schedules asynchronous memory transfers between DRAM and L2, or L2 and L1, to orchestrate the execution of the entire network.

All CNN inferences are executed on GAP8 at a frequency of 170 MHz.
Predictions are transferred via UART to the STM32 MCU, which hosts the other parts of the closed-control loop system.
The control loop involves the same four tasks of~\cite{frontnet}: (i) the CNN model pose estimation, the most computationally intensive task, (ii) a Kalman filter to smooth the sequences of poses, (iii) the velocity controller, and (iv) the low-level control for motor actuation and stabilization.
Note that this paper focuses solely on the perceptual task, although it is expected that reducing its latency and error could also enhance the control loop's performance.
\section{Experimental Results}
\label{sec:results}

\subsection{Setup}
For our experiments, we employed the dataset introduced in~\cite{cereda_deep_2023}, which served as the benchmark for evaluating the performance of the reference static models.
It comprises a total of 30.3k images, divided into training, validation, and test sets in the proportions of 70\%, 20\%, and 10\%, respectively.
We used the validation set to select the most accurate pruned auxiliary CNN and build the error map for Aux-HLC, whereas all results are reported on the test set.
To show the generalization of our method, we also test on a second dataset introduced in~\cite{cereda_deep_2023}.
Similarly to the SoA one, this dataset is collected with a real nano-drone in a different laboratory environment and with different subjects.
This dataset comprises 45k images divided into 72\% for training, 18\% for validation, and 10\% for testing. 
We will henceforth refer to these two datasets as the ``Known Dataset'' and the ``Unseen Dataset.''

Our comparison baselines are the three static CNNs ($F^1$, $F^2$, and $M^{1.0}$) since they constitute the SoA for this task.
In the following sections, we refer to the adaptive solution that employs $F^1$ as the small model as D1 and the one that utilizes $F^2$ as D2.

\subsection{Auxiliary Task Exploration}
\begin{figure}
    \centering
    \includegraphics[width=.97\columnwidth]{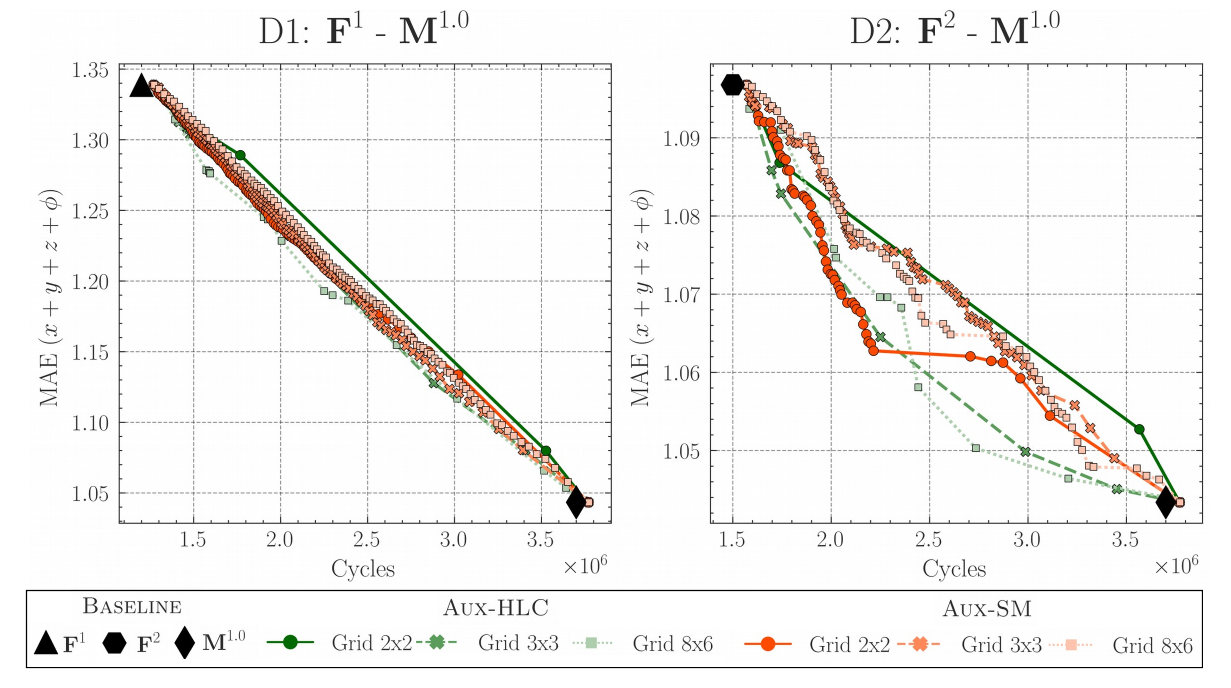}
    \caption{Auxiliary task-based policies comparison on the Known dataset.}
    \label{fig:aux_Manno}
    \vspace{-0.2cm}
\end{figure}

Fig.~\ref{fig:aux_Manno} presents a comprehensive comparison of the auxiliary task policies, considering both CNN pairs and different grid dimensions. 
We did not consider finer grids because doing so would compromise the network's ability to predict the correct quadrant, resulting in nearly random outputs.
The different policies are compared in terms of total MAE on the four regressed variables versus the average number of clock cycles per inference on GAP8, over the Known dataset's test set.
Different points for the same policy refer to different settings of the respective tunable thresholds ($th_{SM}$ and $th_{HLC}$).

While all policies perform similarly on D1, the best results are obtained with Aux-HLC, using an $8\times 6$ grid. 
For D2, the top-2 policies are Aux-HLC with an $8 \times 6$ grid and Aux-SM with a $2 \times 2$ grid, with Aux-HLC achieving optimal performance in lower MAE regions and Aux-SM performing better in the low cycles regime.
For instance, with the Aux-HLC policy, we can obtain a MAE of 1.05, which is only 0.57\% higher than the error of the big model, while simultaneously reducing the inference cycles by 26.07\%. 
In the following sections, we will only consider these two optimal combinations of type and grid size for the Aux policies results.

\subsection{Output vs. Auxiliary Task Policies}

\begin{figure}
    \centering
    \includegraphics[width=.92\columnwidth]{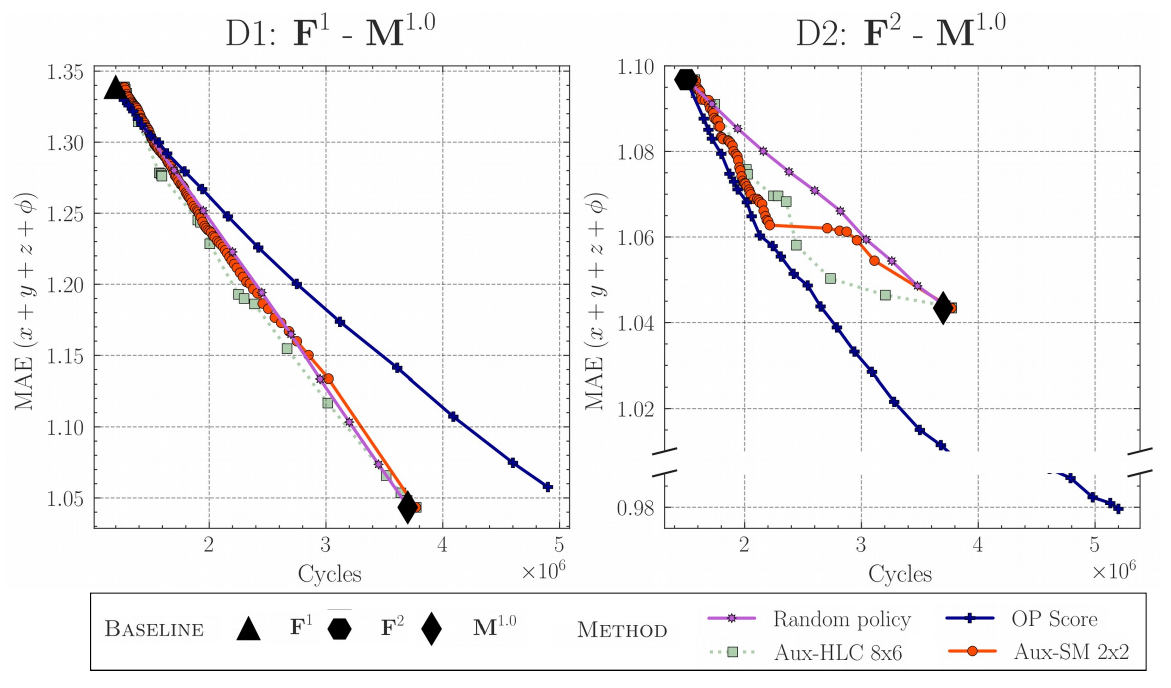}
    \caption{OP and Aux policies comparison on the Known dataset.}
    \label{fig:results_Manno}
    \vspace{-0.2cm}
\end{figure}

In this section, we compare Aux and OP policies on the Known dataset. 
As a baseline for our analysis, we consider a Random policy, that is, a ``trivial'' dynamic model in which a zero-cost policy randomly selects between the execution of the small or the big model.
Fig.~\ref{fig:results_Manno} reports the comparison results for the two adaptive systems, revealing two distinct trends.

For D1, the most effective policy is Aux-HLC.
Interestingly, the OP policy performs worse than the random one in this case.
This can be ascribed to the low accuracy of the small model, which causes a misalignment between the $OP_t$ score and the actual need to invoke the big model. 
The decision to employ the big model becomes then somewhat random. 
This, together with the high cost incurred by the OP policy when the big model is invoked frequently (both models executed for most frames), yields a higher latency without accuracy benefits.

Conversely, for D2, OP significantly outperforms the other policies, resulting in the attainment of new SoA results. 
At iso-MAE with the static big model, we observe a substantial reduction of 28.03\% in inference cycles, whereas at iso-latency, we reduce the MAE by 3.15\%. 
Finally, our approach yields the best known MAE (0.98) for this task, surpassing the current state of the art by 6.13\%.

\subsection{Crazyflie Deployment}

\begin{table}[t]
\centering
\caption{Deployment of different dynamic models and policies on the Crazyflie 2.1.}
\label{tab:deployment_manno}
\renewcommand{\arraystretch}{1.25}
\resizebox{\columnwidth}{!}{%
\begin{tabular}{@{}lllllll@{}}
\hline
\textbf{Models} & \textbf{Method} & \textbf{MAE} & \textbf{Latency} & \textbf{\% Big} & \textbf{Energy} & \textbf{Memory} \\ \hline
\textbf{F$^{1}$} & Static & 1.34 & 7.06 ms & 0 & 0.57 mJ & 153 kB \\
\textbf{F$^{2}$} & Static & 1.10 & 8.82 ms & 0 & 0.71 mJ & 183 kB \\
\textbf{M$^{1.0}$} & Static & 1.04 & 21.76 ms & 100 & 1.92 mJ & 235 kB \\\hline
D1 & Random  & 1.19 & 14.41 ms & 50.0 & 1.25 mJ & 250 kB \\
D1 & Aux-HLC 8x6 & 1.19 & 13.24 ms & 39.1 & 1.14 mJ & 289 kB \\\hline
D2 & Random  & 1.04 & 21.76 ms & 100 & 1.92 mJ & 280 kB \\
D2 & OP & 1.04 & 15.66 ms & 31.4 & 1.32 mJ & 280 kB \\\hline\hline
\end{tabular}%
}
\end{table}

In Tab.~\ref{tab:deployment_manno}, we present a detailed breakdown of some of the Pareto solutions from Fig.~\ref{fig:results_Manno} deployed on the Crazyflie~2.1. 
We select the $th_{*}$ value that maximizes the latency benefit from adaptive inference for each ensemble, comparing the corresponding solutions with the Random policy at iso-MAE.

For the D1 ensemble, the most substantial gain is achieved by the Aux-HLC policy, invoking the big model for 39.1\% of the predictions.
In comparison to the Random Policy, we achieve reductions in latency and energy consumption of 8.1\% and 8.8\%, respectively.
Furthermore, compared to exclusively running the big model, at the cost of a slightly higher total MAE (+0.15), we reduce the latency and energy consumption by 39.1\% and 40.6\%.

For the D2 ensemble, the OP policy significantly outperforms the Random policy. 
The most interesting gains emerge when the adaptive approach is compared to the static big model.
Indeed, running the big model only 31.4\% of the time with our proposed adaptive system is sufficient to maintain the same MAE while concurrently achieving reductions in latency and energy consumption of 28.03\% and 31.25\%, respectively. 

Clearly, all adaptive systems incur a memory overhead since 2 or 3 (in the case of Aux policies) CNNs are deployed on the nano-drone instead of one.
However, as shown in Tab.~\ref{tab:deployment_manno}, even the largest adaptive ensemble easily fits the 512~kB L2 memory of GAP8.
Note that the memory results in the table are computed as the sum of the weights of all the deployed networks and of the biggest activation buffer required to run inference, making the total memory for D1 and D2 less than the sum of the composing static models.

\subsection{Unseen Dataset Results}
\begin{figure}
    \centering
    \includegraphics[width=.95\columnwidth,trim=4 4 4 4,clip]{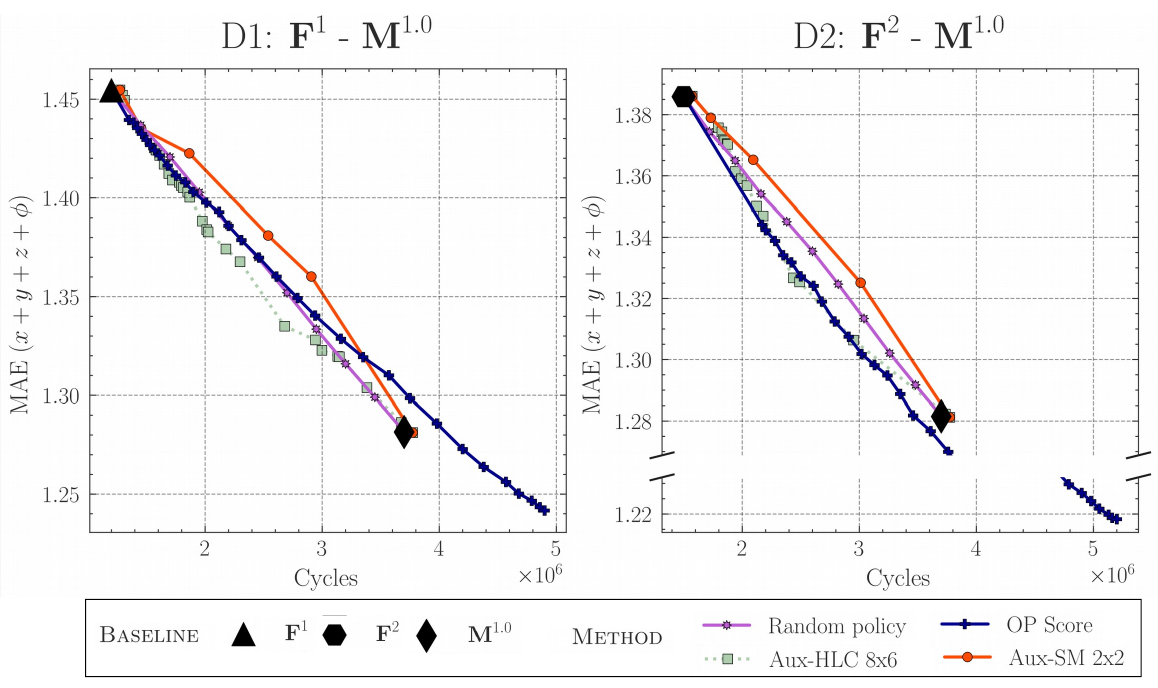}
    \caption{Comparative analysis of the proposed adaptive policies on the Unseen dataset.}
    \label{fig:results_Viganello}
\end{figure}

Finally, Fig.~\ref{fig:results_Viganello} shows the generality of our policies by presenting results on the Unseen dataset. 
The same conclusions drawn from our analysis on the Known Dataset remain valid.
For the D1 ensemble, the best-performing algorithm is Aux-HLC, which achieves the maximum latency reduction (9.2\%) compared to the Random Policy baseline at a MAE of 1.33.
In contrast, for the D2 ensemble, the OP continues to be superior. 
Two significant results it achieves are: (i) the best overall MAE of 1.22, improving the SoA by 4.9\%, and (ii) a 6.49\% latency reduction at iso-MAE with the big model.
\section{Conclusions}
\label{sec:conclusions}

We have presented a practical use case of adaptive inference for perception tasks aboard nano-drones, focusing on visual human pose estimation.
We have derived two adaptive systems from three SoA compact CNNs and proposed and compared different optimal adaptive policies in different working conditions.
These policies have reduced the previous SoA error on the total regressive loss for human pose estimation on nano-drones obtained in~\cite{cereda_deep_2023}. 
On the other hand, fixing the desired MAE to 1.04 (the best result from~\cite{cereda_deep_2023}) and considering the same hardware platform, we achieved a lower latency (15.66 ms) within a power envelope of \SI{90}{\milli\watt}.
Additionally, as shown in Fig.~\ref{fig:results_Manno} and~\ref{fig:results_Viganello}, adaptive inference allows to obtain a rich set of intermediate Pareto-optimal solutions in the latency vs. MAE space, selectable at runtime by changing the value of a single tunable threshold. Future works will include the exploration of more advanced adaptive inference techniques.

\tiny
\bibliographystyle{IEEEtran}
\bibliography{bstctl,references}

\begin{thebibliography}{10}
\providecommand{\url}[1]{#1}
\csname url@samestyle\endcsname
\providecommand{\newblock}{\relax}
\providecommand{\bibinfo}[2]{#2}
\providecommand{\BIBentrySTDinterwordspacing}{\spaceskip=0pt\relax}
\providecommand{\BIBentryALTinterwordstretchfactor}{4}
\providecommand{\BIBentryALTinterwordspacing}{\spaceskip=\fontdimen2\font plus
\BIBentryALTinterwordstretchfactor\fontdimen3\font minus
  \fontdimen4\font\relax}
\providecommand{\BIBforeignlanguage}[2]{{%
\expandafter\ifx\csname l@#1\endcsname\relax
\typeout{** WARNING: IEEEtran.bst: No hyphenation pattern has been}%
\typeout{** loaded for the language `#1'. Using the pattern for}%
\typeout{** the default language instead.}%
\else
\language=\csname l@#1\endcsname
\fi
#2}}
\providecommand{\BIBdecl}{\relax}
\BIBdecl
\renewcommand{\BIBentryALTinterwordstretchfactor}{4}

\bibitem{pulp-dronet}
D.~Palossi \emph{et~al.}, ``An open source and open hardware deep
  learning-powered visual navigation engine for autonomous nano-{UAVs},'' in
  \emph{International Conference on DCOSS}, 2019.

\bibitem{varghese2016challenges}
B.~Varghese \emph{et~al.}, ``Challenges and opportunities in edge computing,''
  in \emph{IEEE International Conference on Smart Cloud}, 2016.

\bibitem{tiny-dronet}
L.~Lamberti \emph{et~al.}, ``Tiny-pulp-dronets: Squeezing neural networks for
  faster and lighter inference on multi-tasking autonomous nano-drones,'' in
  \emph{IEEE International Conference on AICAS}, 2022.

\bibitem{frontnet}
D.~Palossi \emph{et~al.}, ``Fully onboard ai-powered human-drone pose
  estimation on ultralow-power autonomous flying nano-uavs,'' \emph{IEEE IOTJ},
  2022.

\bibitem{cereda_deep_2023}
E.~Cereda \emph{et~al.}, ``Deep neural network architecture search for accurate
  visual pose estimation aboard nano-uavs,'' in \emph{IEEE ICRA}, 2023.

\bibitem{Park2015}
E.~Park \emph{et~al.}, ``{Big/little deep neural network for ultra low power
  inference},'' in \emph{International Conference on CODES+ISSS}, 2015.

\bibitem{Sun_2018_ECCV}
X.~Sun \emph{et~al.}, ``Integral human pose regression,'' in \emph{IEEE ECCV},
  2018.

\bibitem{luvizon18multitask}
D.~C. Luvizon \emph{et~al.}, ``2d/3d pose estimation and action recognition
  using multitask deep learning,'' in \emph{IEEE CVPR}, 2018.

\bibitem{dense-pose}
R.~A. Güler \emph{et~al.}, ``Densepose: Dense human pose estimation in the
  wild,'' in \emph{IEEE CVPR}, 2018.

\bibitem{neuralswarm}
G.~Shi \emph{et~al.}, ``Neural-swarm: Decentralized close-proximity multirotor
  control using learned interactions,'' in \emph{IEEE ICRA}, 2020.

\bibitem{uwbbias}
W.~Zhao \emph{et~al.}, ``Learning-based bias correction for time difference of
  arrival ultra-wideband localization of resource-constrained mobile robots,''
  \emph{IEEE Robotics and Automation Letters}, 2021.

\bibitem{Sze2017}
V.~Sze \emph{et~al.}, ``Efficient {{Processing}} of {{Deep Neural Networks}}:
  {{A Tutorial}} and {{Survey}},'' \emph{Proceedings of the IEEE}, 2017.

\bibitem{pagliari_plinio_2023}
D.~J. Pagliari \emph{et~al.}, ``{PLiNIO}: A user-friendly library of
  gradient-based methods for complexity-aware dnn optimization,'' in
  \emph{{Forum} on {Specification} \& {Design} {Languages} ({FDL})}, 2023.

\bibitem{Daghero2021energy}
F.~Daghero \emph{et~al.}, ``Energy-efficient deep learning inference on edge
  devices,'' in \emph{Hardware Accelerator Systems for Artificial Intelligence
  and Machine Learning}.\hskip 1em plus 0.5em minus 0.4em\relax Elsevier, 2021.

\bibitem{Tann2016}
H.~Tann \emph{et~al.}, ``{Runtime configurable deep neural networks for
  energy-accuracy trade-off},'' in \emph{IEEE CODES}, 2016.

\bibitem{har_journal}
F.~Daghero \emph{et~al.}, ``Human activity recognition on microcontrollers with
  quantized and adaptive deep neural networks,'' \emph{ACM Trans. Embed.
  Comput. Syst.}, 2022.

\bibitem{teerapittayanonBranchyNet2016}
S.~Teerapittayanon \emph{et~al.}, ``{{BranchyNet}}: {{Fast}} inference via
  early exiting from deep neural networks,'' in \emph{ICPR}, 2016.

\bibitem{flamand2018gap}
E.~Flamand \emph{et~al.}, ``Gap-8: A risc-v soc for ai at the edge of the
  iot,'' in \emph{IEEE Int. Conf. on ASAP}, 2018.

\bibitem{conti2017iot}
F.~Conti \emph{et~al.}, ``An iot endpoint system-on-chip for secure and
  energy-efficient near-sensor analytics,'' \emph{IEEE TCAS-I}, 2017.

\bibitem{burrello2020dory}
A.~{Burrello} \emph{et~al.}, ``Dory: Automatic end-to-end deployment of
  real-world dnns on low-cost iot mcus,'' \emph{IEEE Trans. Comput.}, 2021.

\end{thebibliography}

\end{document}